# Fully Convolutional Networks for Text Classification

**Jacob Anderson**
Sentim LLC
Columbus, OH, USA
`papers@sentimllc.com`

## Abstract

**English.** In this work I propose a new way of using fully convolutional networks for classification while allowing for input of any size. I additionally propose two modifications on the idea of attention and the benefits and detriments of using the modifications. Finally, I show suboptimal results on the ITAmoji 2018 tweet to emoji task and provide a discussion about why that might be the case as well as a proposed fix to further improve results.

**Italian.** In questo lavoro viene presentato un nuovo approccio all'uso di fully convolutional network per la classificazione, adattabile a dati di input di qualsiasi dimensione. In aggiunta vengono proposte due modifiche basate sull'uso di meccanismi di attention, valutandone benefici e svantaggi. Infine, sono presentati i risultati della partecipazione al Task ITAmoji 2018 relativo alla predizione di emoji associate al testo di tweets, discutendo il perché delle performance non ottimali del sistema sviluppato e proponendo possibili migliorie.

## 1 Introduction

The dominant approach in many natural language tasks is to use recurrent neural networks or convolutional neural networks (CNN) (Conneau et al., 2017). For classification tasks, recurrent neural networks have a natural advantage because of their ability to take in any size input and output a fixed size output. This ability allows for greater generalization as no data is removed nor added in order for the inputs to match in length. While convolutional neural networks can also support input of any size, they lack the ability to generate a fixed size output from any sized input. In text classification tasks, this often means that the input is fixed in size in order for the output to also have a fixed size.

Other recent work in language understanding and translation uses a concept called attention. Attention is particularly useful for language understanding tasks as it creates a mechanism for relating different position of a single sequence to each other (Vaswani et al., 2017).

In this work I propose a new way of using fully convolutional networks for classification to allow for any sized input length without adding or removing data. I also propose two modifications on attention and then discuss the benefits and detriments of using the modified versions as compared to the unmodified version.

## 2 Model Description

The overall architecture of my fully convolutional network design is shown in Figure 1. My model begins with a character embedding where each character in the input maps to a vector of size 16. I then first apply a causal convolution with 128 filters of size 3. After which, I apply a stack of 9 layers of residual dilated convolutions with skip connections, each of which use 128 filters of size 7. The size of 7 here was chosen by inspection, as it converged faster than size 3 or 5 while not consuming too much memory. Additionally, the dilation rate of each layer of the stack doubles for every layer, so the first layer has rate 1, then the second layer has rate 2, then rate 4, and so on.

All of the skip connections are combined with a summation immediately followed by a ReLU to increase nonlinearity. Finally, the output of the network was computed using a convolution with 25 filters each of size 1, followed by a global max pool operation. The global max pool operation reduces the 3D tensor of size (batch size, input length, 25) to (batch size, 25) in order to match the expected output.

I implemented all code using a combination of Tensorflow (Abadi et al., 2016) and Keras (Chollet, 2015). During training I used softmax cross-entropy loss with an l2 regularization penalty with a scale of .0001. I further reduced overfitting by adding spatial dropout (Tompson et al., 2015) with a drop probability of 10% in the residual dilated convolution layers.

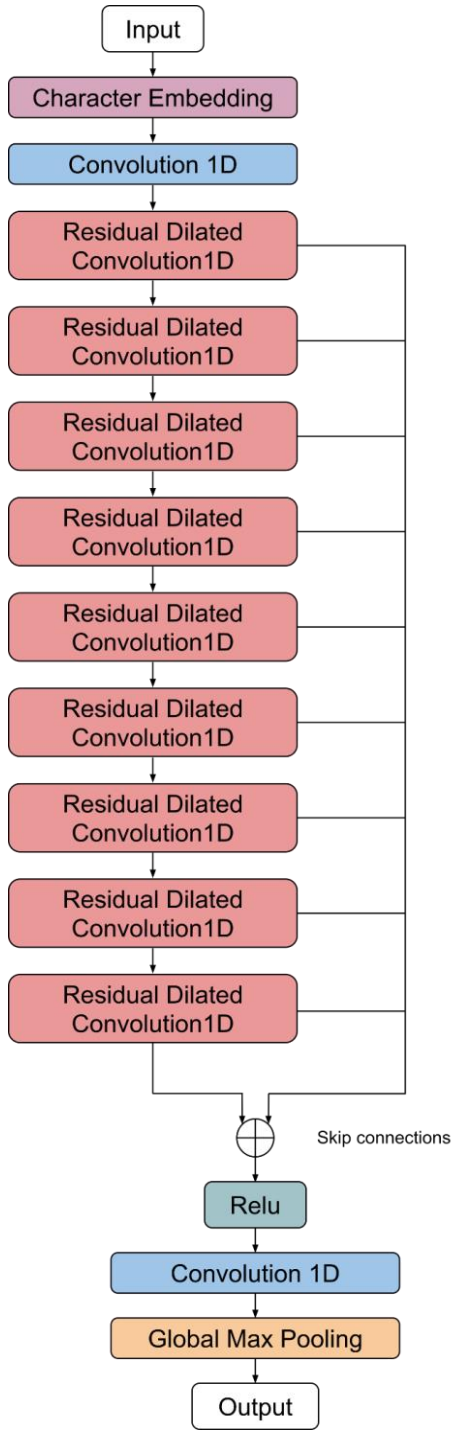

Figure 1: Model Architecture

## 2.1 Hardware Limitations

At the time of creating the models in this paper, I was limited to only a Google Colab GPU, which comes with a runtime restriction of 12 hours per day and a half a GB of GPU memory[1]. While it is possible to continue training again after the restriction is reset, in order to maximize GPU usage, I tried to design each iteration of the model so that it would finish training within a 12 hour time period.

## 2.2 Residual Block

A residual connection is any connection which maps the input of one layer to the output of a layer further down in the network. Residual connections decrease training error, increase accuracy, and increase training speed (He et al., 2015).

## 2.3 Dilated Convolution

A dilated convolution is a convolution where the filter is applied over a larger area by skipping input values according to a dilation rate. This rate usually exponentially scales with the numbers of layers of the network, so you would look at every input for the first layer and then every other input for the second, and then every fourth and so on (van den Oord, 2016).

In this paper, I use dilated convolutions similar to Wavenet (van den Oord, 2016), where each convolution has both residual and skip connections. However, instead of the gated activation function from the Wavenet paper, I used local response normalization followed by a ReLU function. This activation function was proposed by Krizhevsky, Sutskever, and Hinton (2012), and I used it because I found this method to achieve equal results but faster convergence.

## 2.4 Residual Dilated Convolution

A residual dilated convolution is a dilated convolution with a residual connection. First, I take a dilated convolution on the input and a linear projection on the input. The dilated convolution and the linear projection are added together and then outputted. The dilated convolution also outputs as a skip connection, which is eventually summed together with every other skip connection later in the network.

---

[1] They have since changed this limitation to 13 GB.

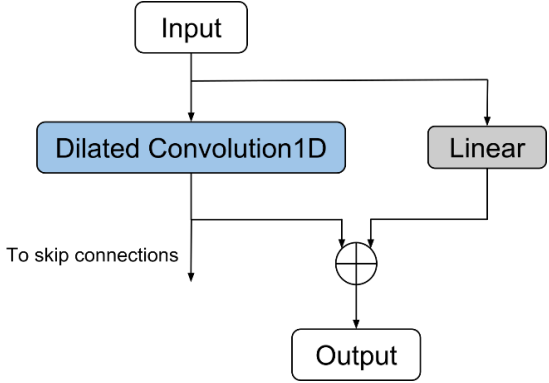

Figure 2: Residual Dilated Convolution

## 2.5 Skip Connections

In this paper, I also use the idea of skip connections from Long, Shelhamer, and Darrell (2015). Skip connections simply connect previous layers with the layer right before the output in order to fuse local and global information from across the network. In this work, the connections are all fused together with a summation followed by a ReLU activation to increase nonlinearity.

## 2.6 Attention and Self-Attention

Attention can be described as mapping a query and a set of key value pairs to an output (Vaswani et al., 2017). Specifically, when I say attention or 'normal' attention, I am referring to Scaled Dot-Product Attention. Scaled Dot-Product Attention is computed as:

$$Attention(Q, K, V) = softmax\left(\frac{QK^T}{\sqrt{d_k}}\right)V \quad (1)$$

Where Q, K, and V are matrices representing the queries, keys, and values respectively (Vaswani et al., 2017).

Self-Attention then is where Q, K, and V all come from the same source vector after a linear projection. This allows each position in the vector to attend to every other position in the same vector.

## 2.7 Simplified and Local Attention

Simplified and local attention can both be thought of as trying to reinforce the mapping of a key to value pair by extracting extra information from the key. I compute a linear transformation followed by a softmax to get the weights on the values. These weights and the initial values are multiplied together element-wise in order to highlight which of the values are the most important for solving the problem. Simplified attention can also be thought of as reinforcing a one-to-one correspondence between the key and the value.

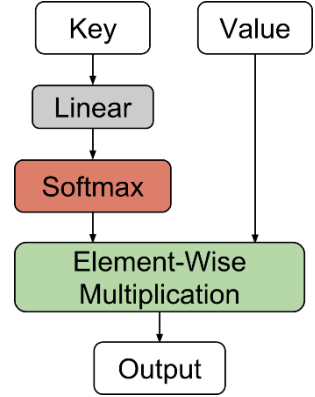

Figure 3: Simplified Attention

Local attention is like simplified attention except instead of performing a linear projection on the keys, local attention performs a convolutional projection on the keys. This allows for the network to use local information in the keys to attend to the values.

## 2.8 Multi-Head Attention

In multi-head attention, attention is performed multiple times on different projections of the input (Vaswani et al., 2017). In this paper, I either use one or eight heads in every experiment with attention, in order to get the best results and to compare the different methods accurately.

## 2.9 Model Modifications for Attention

In this paper, I tested seven different models, six of which extend the base model using some type of attention. In the models with attention, self-attention is used right after the final convolution and right before the global pooling operation.

## 2.10 Global Max Pooling

While CNN's support input of any size, they lack the ability to generate a fixed size input and instead output a tensor that is proportional in size to the input size. In order for the output of the network to have a fixed size of 25, I use max pooling (Scherer et al., 2010) along the time dimension of the last convolutional layer. I perform the max pooling globally, which simply means that I take the maximum value of the whole time dimension instead of from a sliding window of the time dimension.

## 3 Experiment and Results

In this section, I go over the ITAmoji task description and limitations, as well as my results on the task.

### 3.1 ITAmoji Task

This model was initially designed for the ITAmoji task in EVALITA 2018 (Ronzano et al., 2018). The goal of this task is to predict which of 25 emojis (shown in Table 1) is most likely to be in a given Italian tweet. The provided dataset is 250,000 Italian tweets with one emoji label per tweet, and no additional data is allowed for training the models. However, it is allowed to use additional data to train unsupervised systems like word embeddings. All results in the coming subsections were tested on the dataset of 25,000 Italian tweets provided by the organizers.

| Emoji | Label | % Samples |
|---|---|---|
|  | red heart | 20.28 |
|  | face with tears of joy | 19.86 |
|  | smiling face with heart eyes | 9.45 |
|  | kiss mark | 1.12 |
|  | winking face | 5.35 |
|  | smiling face with smiling eyes | 5.13 |
|  | beaming face with smiling eyes | 4.11 |
|  | grinning face | 3.54 |
|  | face blowing a kiss | 3.34 |
|  | smiling face with sunglasses | 2.80 |
|  | thumbs up | 2.57 |
|  | rolling on the floor laughing | 2.18 |
|  | thinking face | 2.16 |
|  | blue heart | 2.02 |
|  | winking face with tongue | 1.93 |
|  | face screaming in fear | 1.78 |
|  | flexed biceps | 1.67 |
|  | face savoring food | 1.55 |
|  | grinning face with sweat | 1.52 |
|  | loudly crying face | 1.49 |
|  | top arrow | 1.39 |
|  | two hearts | 1.36 |
|  | sun | 1.28 |
|  | rose | 1.06 |
|  | sparkles | 1.06 |

Table 1: Each of the 25 different emojis used in the ITAmoji task, their labels, and the corresponding percent of samples in the test dataset.

### 3.2 Results

Table 2 shows my official results from the ITAmoji competition, as well as the first and second group scores. Table 3 shows the best result (evaluated after the competition was complete) according to the macro f1 score of the seven different models I trained during the competition. It also shows the micro f1 score at the same run of the best macro f1 score for comparison. Table 4 shows the upper and lower bounds of the f1 scores after the scores have stopped increasing and have plateaued.

| Model | Macro F1 | Micro F1 |
|---|---|---|
| 1st Place Group | 0.365 | 0.477 |
| 2nd Place Group | 0.232 | 0.401 |
| Run 3: Simplified Attention | 0.106 | 0.294 |
| Run 2: 1 Head Attention | 0.102 | 0.313 |
| Run 1: No Attention[2] | 0.019 | 0.064 |

Table 2: Official results from the ITAmoji competition, as compared to the first and second place groups.

| Model | Macro F1 | Micro F1 |
|---|---|---|
| 8 Head Attention | 0.113 | 0.316 |
| 1 Head Attention | 0.105 | 0.339 |
| Local Attention | 0.106 | 0.341 |
| 8 Head Local | 0.106 | 0.337 |
| Simplified Attention | 0.106 | 0.341 |
| 8 Head Simplified | 0.109 | 0.308 |
| No Attention | 0.11 | 0.319 |

Table 3: The best results from the different models on the dataset, run after the competition was over.

---

[2] Due to an off-by-one error in the conversion from network output to emoji, the official results for the no attention network are much worse than in actuality.

| Model | Macro F1 | Micro F1 |
|---|---|---|
| 8 Head Attention | [.10, .11] | [.30, .36] |
| 1 Head Attention | [.09, .11] | [.30, .36] |
| Local Attention | [.10, .11] | [.30, .35] |
| 8 Head Local | [.10, .11] | [.34, .36] |
| Simplified Attention | [.10, .11] | [.32, .36] |
| 8 Head Simplified | [.10, .11] | [.31, .36] |
| No Attention | [.10, .11] | [.30, .36] |

Table 4: The upper and lower bounds of the f1 scores of the different model types after the scores have plateaued in training and start oscillating.

While 8 head attention did outperform the 8 head local and simplified models, it's interesting to note that that isn't the case for the 1 head versions. Additionally, the bounds for the scores significantly overlap so there is no statistically significant gains for one method over the other. This result, along with my comparatively worse scores is probably because the max pooling at the end of my model was throwing away too much information in order to make the size consistent.

## 4 Discussion

In the upcoming sections, I discuss a possible problem with the design of my models and propose a few solutions for that problem. I further discuss the two new modifications on attention that I proposed and their possible uses.

### 4.1 Loss of Information While Pooling

For the problem of throwing away too much information during the pooling or downsampling phase, there are three main approaches that could be explored, each with their positives and negatives.

The first approach is to just fix the size of the input and use fully connected layers or similar approaches to find the correct output. This is the current approach by most researchers, and has shown good results. The main negative here is that the input size must be fixed, and fixing the input size could mean throwing away or adding information that isn't naturally there.

The second approach is to use a recurrent neural network neuron like an LSTM or a GRU with size equal to the output size to parse the result and output singular values for the final sequence. This would probably lead to better results but is going to be slower than the other approaches.

The last approach is to use convolutional layers with a large kernel size and stride (e.g. stride equal to the size of the kernel). This would allow the network to shrink the output size naturally, and would be faster than using an LSTM. The issue here is that in order to maintain the property that the network can have any input size, pooling or some other method of downsampling has to be used, potentially throwing away useful data.

### 4.2 Potential Uses of Simplified and Local Attention

While the original idea behind simplifying attention in such a manner as presented in this paper was to reduce computational cost and encourage easier learning by enforcing a softmax distribution of data, there didn't seem to be any benefit in doing so. In most cases the computational cost of a couple of matrix multiplications versus an element-wise product is negligible, so it would usually be better to just apply normal attention in those cases as it already covers the case of simplified attention in its implementation.

Similar to simplified attention, it doesn't necessarily make sense to use local attention instead of normal attention for small input sizes. Instead, it might make sense to switch out the linear projection on the queries and keys in normal attention with a convolutional projection but otherwise perform the scaled-dot product attention normally. This could be potentially useful if the problem being approached needs to map patterns to values instead of mapping values to values. One could of course extend this even further by also performing a convolutional projection on the values in order to map local patterns to other local patterns, and so on.

On the other hand, the local attention suggested in this paper could be useful in neural nets used for images and other large data, where it might not make sense to attend over the whole input. This is especially true in the initial layers of such neural networks where the neurons are only looking at a small section of the input in the first place. Beyond the smaller memory demands compared to normal attention, local attention could be useful in these layers because it provides a method to naturally figure out which patterns are important at these early layers.

Of course an alternative to local attention is to just take small patches of the image and apply the original formulation of scaled-dot product attention to get similar results. This idea was originally suggested as future work in Vaswani et al. (2017).

## 5 Conclusion

In this work I present simplified and local attention and test the methods in comparison to similar models with normal attention and without any kind of attention at all. I also introduced a new strategy for classifying data with fully convolutional networks with any sized input.

The new model design was not without its own flaws, as it showed poor results for all modifications of the method. The poor results were probably due to the final pooling layer throwing away too much information. A better method would be to use LSTMs or specially designed convolutions in order to shrink the output to the correct size.

Future work will include further explorations of simplified and local attention to really get a grasp of what tasks they are good at and where, if anywhere, they show better efficiency or results than normal attention. In the future I will also further explore the new strategy for classification on any sized input with fully convolutional model and see what I can change and update in order to improve the results of the model.